\documentclass[letterpaper, 10 pt, conference]{ieeeconf}
\IEEEoverridecommandlockouts
\let\labelindent\relax
\usepackage{enumitem}
\usepackage{balance}
\usepackage[font={small}]{caption}
\usepackage[labelformat=simple]{subfig}
\usepackage{array}
\usepackage{textcomp}
\usepackage{mathtools, nccmath}
\usepackage{graphicx}
\usepackage{amsfonts}
\usepackage{amsmath}
\usepackage{autobreak}
\usepackage{amssymb}
\usepackage{tikz}
\usepackage{arydshln}
\usepackage{multirow}
\usepackage{bm}
\usepackage{epstopdf}
\usepackage{cite}
\usepackage{siunitx}
\usepackage{xcolor}
\usepackage{float} 
\usepackage{stfloats}
\usepackage{lineno}
\usepackage{subfloat}
\usepackage{colortbl}
\usepackage{booktabs}
\usepackage[ruled,linesnumbered,boxed,noend]{algorithm2e}
\usepackage{float}
\makeatletter
\let\NAT@parse\undefined
\makeatother
\usepackage[colorlinks=true,
            linkcolor=blue,
           ]{hyperref}

\allowdisplaybreaks[4]

\title{\LARGE \bf CDF-Glove: A Cable-Driven Force Feedback Glove \\for Dexterous Teleoperation 
}

\author{
Huayue Liang, Ruochong Li, Yaodong Yang, Long Zeng, Yuanpei Chen$^\dagger$ and Xueqian Wang$^\dagger$
\thanks{$^\dagger$ Corresponding authors: Yuanpei Chen and Xueqian Wang.}
\thanks{This work was partially supported by Shenzhen Science and Technology Program (Grant No. JCYJ20220818101014030).}
\thanks{Huayue Liang and Xueqian Wang are with the Center for Artificial Intelligence and Robotics, Shenzhen International Graduate School, Tsinghua University, Shenzhen 518055, China, {\tt\footnotesize\{lianghy23@mails, wang.xq@sz\}.tsinghua.edu.cn}.}
\thanks{Yuanpei Chen, Ruochong Li, and Yaodong Yang are with the PKU-Psibot Joint Lab, Peking 100871, China, \tt {yuanpei.chen312@gmail.com}, {rclihkust@gmail.com}, {yaodong.yang@pku.edu.cn}.}
\thanks{Long Zeng is with the Department of Advanced Manufacturing, Shenzhen International Graduate School, Tsinghua University, Shenzhen 518055, China, 
\tt {zenglong@sz.tsinghua.edu.cn}.}
\thanks{This work was completed during an internship at PsiBot.}
}

\begin{document}

\maketitle
\begin{abstract}
High-quality teleoperated demonstrations are a primary bottleneck for imitation learning (IL) in dexterous manipulation. 
However, haptic feedback provides operators with real-time contact information, enabling real-time finger-posture adjustments, and thereby improving demonstration quality.
Existing dexterous teleoperation platforms typically omit haptic feedback and remain bulky and expensive.
We introduce CDF-Glove, a lightweight and low-cost cable-driven force-feedback glove.
The real-time state is available for 20 finger degrees of freedom (DoF), of which 16 are directly sensed and 4 are passively coupled (inferred from kinematic constraints).
We develop a kinematic model and control stack for the glove, and validate them across multiple robotic hands with diverse kinematics and DoF.
The CDF-Glove achieves distal-joint repeatability of 0.4°, and delivers about 200 ms force-feedback latency, yielding a 4× improvement in task success rate relative to no-feedback teleoperation.
We collect two bimanual teleoperation datasets, on which we train and evaluate Diffusion Policy baselines.
Compared to kinesthetic teaching, the policies trained in our teleoperated demonstrations increase the average success rate by 55\% and reduce the mean completion time by $\approx$ 15.2 seconds (a 47.2\% relative reduction).
In particular, the CDF-Glove costs $\approx$ US\$230.
The code and designs are released as open source at \url{https://cdfglove.github.io/}.

\end{abstract}

\section{Introduction}
\label{sec:Introduction}

Imitation Learning (IL) \cite{ARGALL2009469} has emerged as a powerful paradigm for solving complex dexterous manipulation tasks, yet its success is critically dependent on the availability of high-quality expert demonstration data.
Data gloves~\cite{8206575} present an intuitive interface for capturing human hand kinematics for this purpose. However, the performance of the resulting IL policy is strongly dependent on the quality and dimensionality of the captured data. This motivates the development of a teleoperation system capable of recording high-dimensional hand joint data with high precision.
Furthermore, haptic feedback~\cite{8794230} is critical, as it enables the operator to perceive the haptic interaction during grasping and manipulation. This closed-loop control allows the operator to make nuanced adjustments to their manipulation strategies, thereby significantly enhancing the quality and effectiveness of the collected demonstration data.
Despite its importance, existing high-DoF haptic gloves~\cite{zhang2025doglove,7759176} present a challenging trade-off: they are often prohibitively expensive, limited in sensing dimensionality and actuated DoF, or lack haptic feedback entirely. In addition, many existing systems neglect crucial aspects of usability, such as wearability, comfort, and inherent safety, which limit their practical application for long-duration data collection.
To address these limitations, this work introduces the \textbf{CDF-Glove}: a lightweight, low-cost, and high-DoF haptic glove. Our design leverages a cable-driven mechanism, which offers inherent safety, rapid response times, and a simple structure conducive to easy replication. 
Subsequently, the kinematics of the cable-driven finger measurement system and the tracking of force-feedback cables are modeled and analyzed.
Finally, a series of experiments are conducted to verify the performance of the CDF-Glove. Bimanual data collection is performed using the CDF-Glove. The data collected are used for IL training to verify both the high quality of the data and the effectiveness of the Diffusion Policy~\cite{chi2024diffusionpolicy}. The collected data should be helpful in other dexterous manipulation tasks \cite{JIANG2026103923,10160491}.

\begin{figure}
    \centering
    \includegraphics[width=8.5cm]{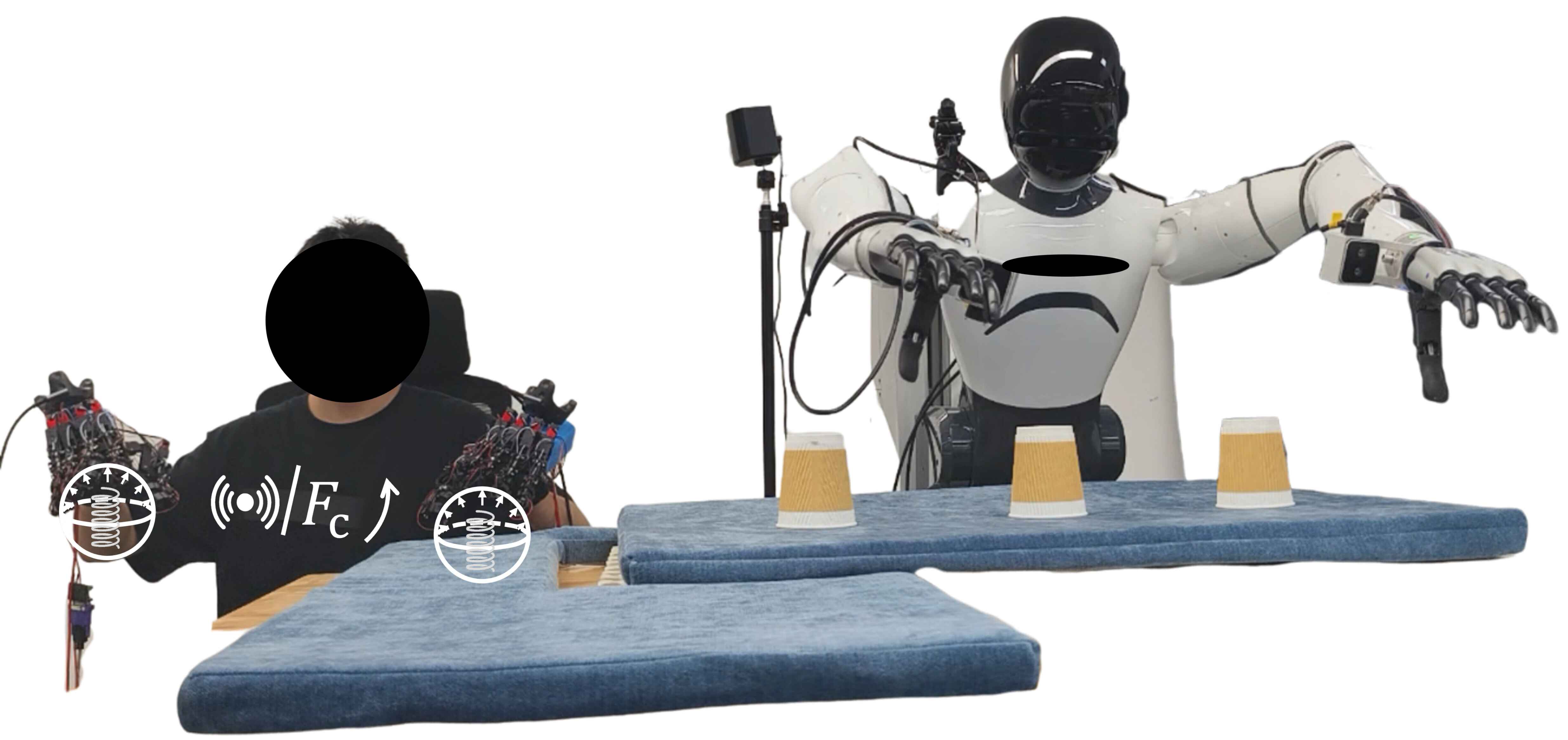}
    \caption{
    The CDF-Glove and the whole machine teleoperation system. The CDF-Glove is worn on the operator's hand to collect hand joint data and provide haptic feedback. 
    }
    \label{fig:main}
    \vspace{-0.2cm}
\end{figure}

This work offers the following contributions:
    \begin{itemize}
        \item The design and implementation of the CDF-Glove, a novel, lightweight, high-DoF, and inherently safe haptic glove with an integrated haptic feedback system. The device achieves a repeatability of the position of the distal joint of 0.4° and a response time to haptic feedback of 200 ms. The CDF-Glove costs $\approx$ \$230.
        \item A comprehensive kinematic model that maps cable displacement to finger joint angles, alongside a tracking model for the force-feedback transmission system.
        \item An extensive experimental validation demonstrating that IL policies trained with data collected by the CDF-Glove significantly outperform those from kinesthetic teaching (KT). Specifically, our method improves the task success rate by 55\% percentage points and reduces the average completion time by $\approx$ 15.2 seconds (a 47.2\% relative reduction).
    \end{itemize}
\section{related work}
\label{sec:related work}
    
\subsection{General Demonstration Teleoperation}
\label{General Demonstration Teleoperation}
In the field of teleoperation, various data collection methods have been explored. 

The VR-based approach\cite{qin2023anyteleop,Zheng2025DemonstratingDD} is simple and easy to deploy, making it suitable for rapid development and testing. However, it suffers from poor dynamic performance, especially during fast movements or complex tasks, leading to latency and data loss. In addition, it lacks precision in mapping the hand posture, which is crucial for high-precision dexterous operations.

The motion capture approach using cameras\cite{wang2024dexcap,naughton2024respilotteleoperatedfingergaiting} provides intuitive operation and can capture overall movements well. However, it falls short of accurately mapping hand movements, especially the fine details of finger actions. Its dynamic performance is average, and the high cost of motion-capture gloves limits its practical application.

Despite their generalization potential, existing methods often incur high costs and tracking latencies, hindering operational smoothness. Furthermore, the lack of haptic feedback and insufficient precision for fine-grained movements make it challenging for operators to accurately adjust strategies during complex tasks.

\subsection{Wearable Devices for Teleoperation}
\label{General Demonstration Teleoperation} 

Many studies have also utilized wearable devices for data collection in teleoperation.

The combination of exoskeleton arms with grippers or exoskeleton hands\cite{ben2024homie,zhao2023learningfinegrainedbimanualmanipulation,fang2025airexo2scalinggeneralizablerobotic} offers lower hardware costs but results in a more complex system that is difficult to maintain. These systems are often bulky, restricting the operator's range of motion and reducing operational flexibility. Moreover, the performance of their interaction, particularly in haptic feedback, is not satisfactory, leading to an unnatural operator experience. On the other hand, linkage exoskeletons  \cite{doi:10.1126/scirobotics.adn3802,6569265} are mainly bulky and heavy, with complex wearable procedures that can negatively impact operator experience, resulting in low-quality data collection. Furthermore, the kinematic calculations for these exoskeletons are complicated. 

Furthermore, the cost of Zhang et al.'s \cite{zhang2025doglove} exoskeleton is still not low enough, and its significant weight can affect the operator's experience and the precision of data. Meanwhile, many cable-driven exoskeletons \cite{8918345,9090275} lack accurate position sensing or have complex control mechanisms, resulting in limited data collection and low DoF.

To overcome these issues, we propose CDF-Glove: a lightweight, low-cost, and high-DoF teleoperation system. By integrating trackers and haptic feedback, it provides a viable, cost-effective solution for collecting the diverse human data. This is crucial for advanced learning frameworks like DexGraspVLA\cite{zhong2025dexgraspvla}, which leverage foundation models for robust dexterous manipulation.

\section{mechanical design}
\label{sec:mechanical design}

\begin{figure*}[t]
    \centering
    \includegraphics[width=\linewidth]{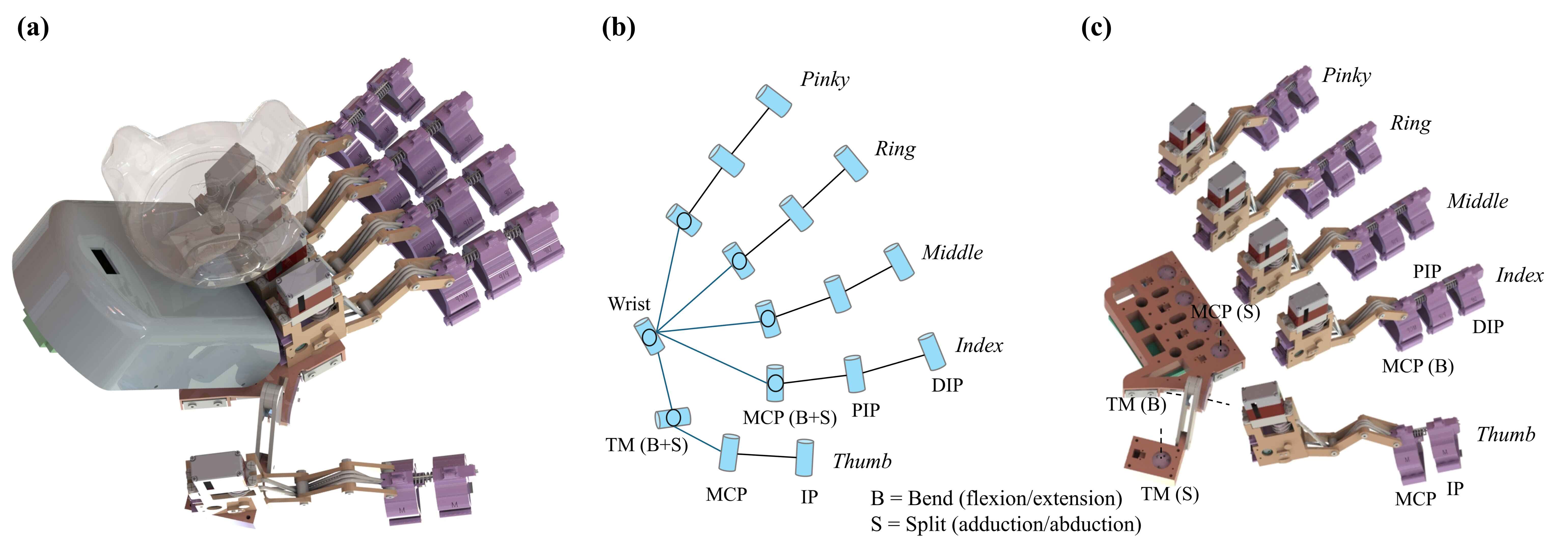}
    \caption{
Overview of the CDF-Glove system. (a) Overall structural diagram of the CDF-Glove system. (b) Kinematic structure of the human hand system. (c) Overall kinematic layout of the CDF-Glove system.
    }
    \label{fig:zongtu_sanzhang}
    \vspace{-0.3cm}
\end{figure*}

\begin{figure}[t]
    \centering
    \includegraphics[width=\linewidth]{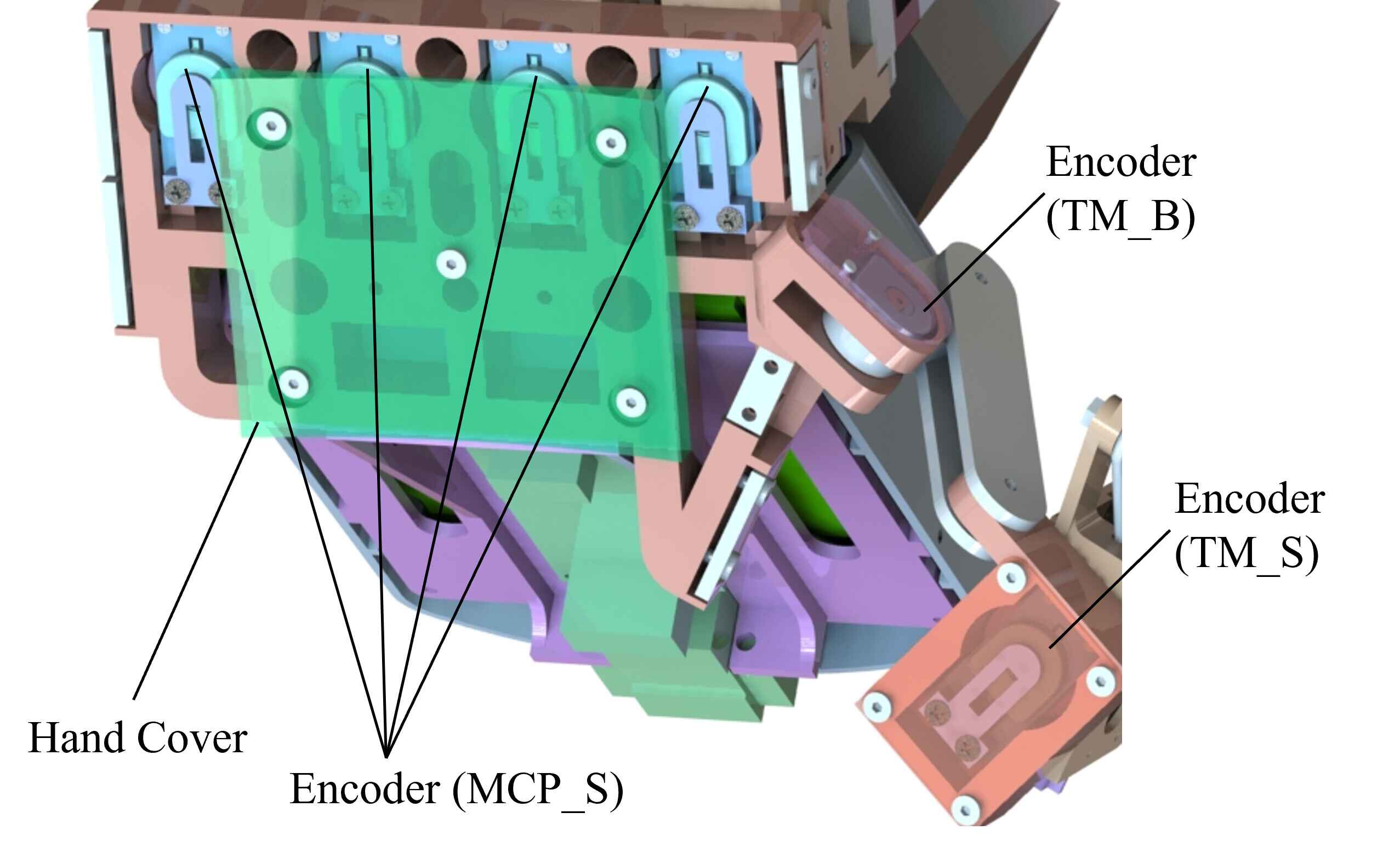}
    \caption{
Arrangement of encoders at the base of the fingers.
    }
    \label{fig:dianweiqitu}
    \vspace{-0.3cm}
\end{figure}

\begin{figure}[t]
    \centering
    \includegraphics[width=8.65cm]{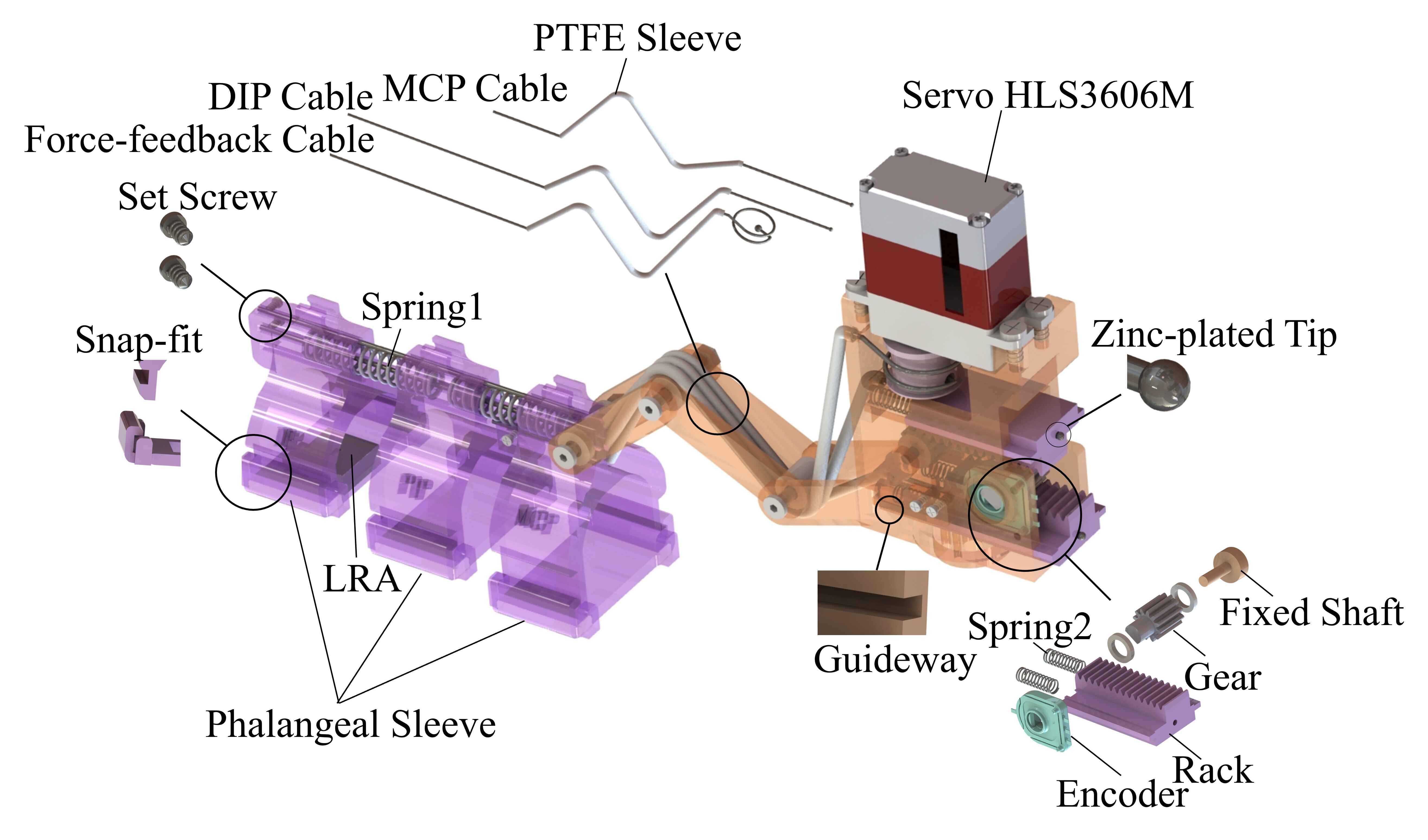}
    \caption{
    Exploded view of the single-finger structure.
    }
    \label{fig:baozatu}
    \vspace{-0.3cm}
\end{figure}

\subsection{Design Requirements and Approach}
\label{subsec:Design Requirements and Approach}
CDF-Glove is a lightweight, compact wearable designed for high-fidelity hand motion capture and operator comfort. To ensure reliability, it employs a closed-loop cable-driven mechanism that provides real-time haptic feedback while consolidating components on the back of the hand to eliminate bulky linkages. This architecture leverages the low latency of cable transmission for real-time performance and enhances safety by automatically relaxing the cables during abnormal force detection to prevent injury.

Given the characteristics of the CDF-Glove, it is divided into a measurement system in Sec. \ref{subsec:Measurement System} and a force feedback system in Sec. \ref{subsec:Haptic and Force Feedback System}.

\subsection{Measurement System}
\label{subsec:Measurement System}
Many studies \cite{CERULO201775} have modeled the kinematics of the human hand. As shown in Fig. \ref{fig:zongtu_sanzhang}(b), the hand consists mainly of hinge joints and ball-and-socket joints. 
Each index, middle, ring, and pinky fingers features two 1-DoF hinge joints—the distal interphalangeal (DIP) and proximal interphalangeal (PIP)—that perform bend. Their metacarpophalangeal (MCP) joint is a 2-DoF joint, enabling both bend and split. 
The thumb differs slightly: both the IP and MCP joints allow only bending, whereas the trapeziometacarpal (TM) joint is a saddle joint that allows both bending and split motions. Additionally, the wrist provides two DoF. 

Based on these anatomical features and joint arrangements, we designed an exoskeleton glove system, CDF-Glove, as illustrated in Fig. \ref{fig:zongtu_sanzhang}(a). The wrist is tracked using the \textit{HTC} VIVE Tracker 3.0, which is shown semi-transparently in Fig. \ref{fig:zongtu_sanzhang}(a). After rigidly attaching the hand cover shown in Fig. \ref{fig:dianweiqitu} to the glove, the assembly is worn on the operator's hand. It integrates 16 active measured DoF and 4 passive DoF (PIP).
As depicted in Fig. \ref{fig:zongtu_sanzhang}(c), the kinematic model of each hand joint is mapped onto the CDF-Glove. Joint data are acquired using RDC506018A encoders. For the index, middle, ring, and pinky fingers, the split motion of the MCP joint is directly measured by encoders, as shown in Fig. \ref{fig:dianweiqitu}. For the thumb, both DoF of the TM joint are also measured by encoders.

The structure of each finger, depicted in Fig. \ref{fig:baozatu}, consists mainly of non-standard components 3D printed from PLA for a lightweight and low-cost design. Each phalanx is equipped with a dedicated sleeve, which attaches securely to the finger via a snap-fit mechanism to prevent slippage and rotation. The adjacent phalangeal sleeves are connected by the connecting spring (spring 1), which not only links the sleeves but also enables faster force feedback reset. The wire diameter of spring 1 is small, resulting in low stiffness that can be neglected during operation, thus minimizing interference with natural hand movements.

To measure the angles of the DIP and MCP joints, two steel cables with zinc-plated tips and a diameter of 1 mm are routed through the DIP and MCP joints, respectively. These cables are guided by polytetrafluoroethylene (PTFE) sleeves to effectively reduce friction \cite{RADULOVIC20241001}. 
One end of each cable passes through the reset spring (spring 2) and is secured to a rack by the zinc-plated tip. The rack slides along a guideway, driving a 12-tooth gear to rotate, thereby changing the encoder readings. 
The moderate stiffness of spring 2 avoids significant interference with hand operation, ensures encoder reset, and effectively prevents cable slack during tensioning, thus maintaining measurement accuracy. The opposite end of the steel cable is fixed to the phalangeal sleeve using a set screw. After an initial pre-tensioning, the system does not require further adjustment during an operational session. Specifically, the MCP and DIP cables are fixed to their phalangeal sleeves, allowing independent measurement of the DIP and MCP joint angles. 
The calculation method for the PIP angle will be described in Sec. \ref{subsec:encoder jisuan formula}.

\subsection{Force Feedback System}
\label{subsec:Haptic and Force Feedback System}
Our device integrates a dual-mode haptic feedback system, combining vibrotactile feedback from Linear Resonant Actuators (LRAs) with kinesthetic force feedback from a cable-driven mechanism, as illustrated in Fig. \ref{fig:baozatu}, which together constitute the haptic and cable-driven force feedback system. 
The system employs two distinct types of actuators: LRAs provide vibrotactile feedback directly to the operator's fingertip, while servos drive a cable-transmission system to generate kinesthetic force feedback.
When the dexterous hand equipped with force sensors transmits signals to the main control board, the force feedback mechanism is activated, as summarized in Table. \ref{tab:feedback}.

The LRA is embedded beneath the DIP phalangeal sleeve, allowing direct contact with the operator's fingertip to deliver high-frequency haptic feedback. Depending on the measured force, the LRA automatically adjusts the frequency of its driving signal, producing different waveforms to achieve various haptic feedback effects. As shown in Table. \ref{tab:feedback}, the LRA driving signals are divided into two ranges: 0.1\url{~}0.5\,N (Waveform 1) and 0.5\url{~}1\,N (Waveform 2).

The \textit{FEETECH} HLS3606M servo provides a stall torque of 6\,kg$\cdot$cm, which is sufficient to meet the requirements.
One end of the force-feedback cable is fixed to the DIP phalangeal sleeve with a set screw, while the other end is connected to the output flange, passing through the PTFE sleeve. The servo drives the output flange, tightens the cable, and generates force feedback. The tension in the cable can be adjusted by the servo's rotation, enabling different levels of force feedback. When the finger bends, the cable elongates, and the amount of elongation is calculated using the formula described in Sec. \ref{subsec:force jisuan formula}.

\begin{table}[t]
\centering
\caption{Haptic Feedback Strategy of CDF-Glove}
\label{tab:feedback}
\begin{tabular}{@{}ccc@{}}
\toprule
\textbf{Force Sensor Readings (N)} & \textbf{Vibration Feedback} & \textbf{Force Feedback}\\ 
\midrule
$<0.1$   & --- & ---\\
$0.1$\url{~}$0.5$ & \checkmark (Waveform 1) & ---\\
$0.5$\url{~}$1$ & \checkmark (Waveform 2)& ---\\
$>1$ & --- & \checkmark\\
\bottomrule
\end{tabular}
\end{table}

\section{System Modeling}
\label{sec:System Modeling}
\subsection{Kinematic Model of Finger Joint}
\label{subsec:encoder jisuan formula}
\begin{figure}[t]
    \centering
    \includegraphics[width=\linewidth]{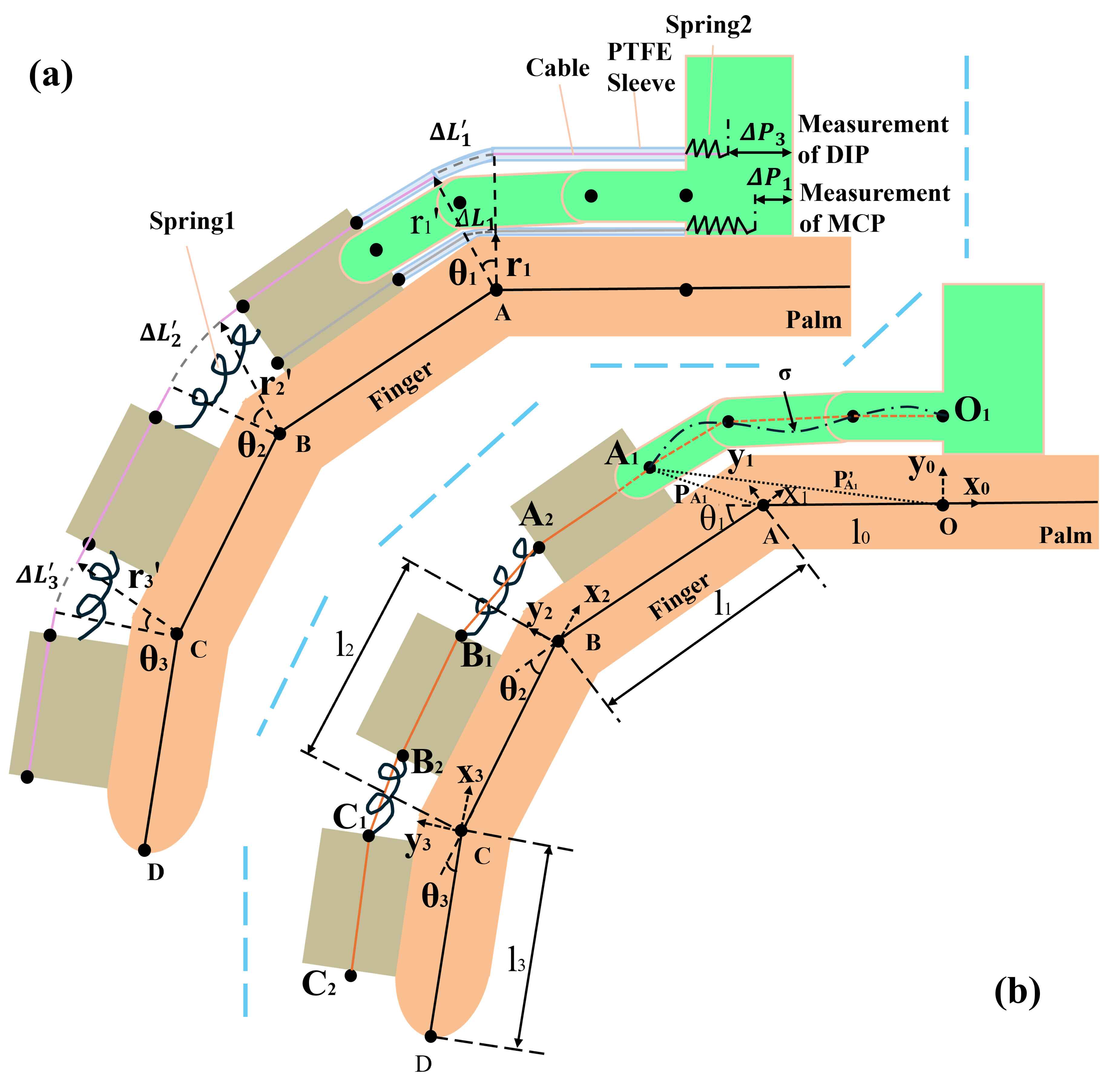}
    \caption{
    Kinematic model for a single finger joint in the flexed position. (a) Schematic of joint angle calculation for an individual finger. (b) Force-feedback cable following calculation based on fingertip position localization.
    }
    \label{fig:shouzhijiesuan}
    \vspace{-0.3cm}
\end{figure}

To retarget motion to a dexterous robotic hand, we first establish a kinematic model of the CDF-Glove. This model computes the operator's finger joint angles from the device's encoder readings.
As shown in Fig. \ref{fig:shouzhijiesuan}(a), only the angles of the MCP and DIP joints are directly measured, while the angle of the PIP joint is correlated with the angle of the DIP joint \cite{6987303}, as reflected in \eqref{equ:Θ2}.
Therefore, the PIP joint angle can be computed from the DIP joint angle. The kinematic model can thus be formulated as follows:

\begin{equation}
\begin{gathered}
\theta_{1} = \frac{\Delta P_{1}}{r_{1}} , \\
\Delta P_{1} = r_{g} \cdot \theta_{e_m} ,\\
\Delta L_{1}' = \theta_{1} \cdot r_{1}'
\end{gathered}
\label{equ:Θ1}
\end{equation}

\begin{equation}
\begin{gathered}
\theta_{2} = \frac{\theta_{3} + 0.230}{0.989} , \\
\Delta L_{2}' = \theta_{2} \cdot r_{2}'
\end{gathered}
\label{equ:Θ2}
\end{equation}

\begin{equation}
\begin{gathered}
\theta_{3} = \frac{\Delta L_{3}'}{r_{3}'} , \\
\Delta P_{3} = r_{g} \cdot \theta_{e_d} , \\
\Delta P_{3} = \Delta L_{1}' + \Delta L_{2}' + \Delta L_{3}'
\end{gathered}
\label{equ:Θ3}
\end{equation}
where $\theta_{1}$, $\theta_{2}$, and $\theta_{3}$ represent the angles of the MCP, PIP, and DIP joints, respectively. $\Delta P_{1}$ and $\Delta P_{3}$ denote the cable displacements at the MCP and DIP joints, respectively. $r_{1}'$, $r_{2}'$, and $r_{3}'$ are the effective radii of the DIP measurement cable at the MCP, PIP, and DIP joints, with values of 23.25\,mm, 19.31\,mm, and 17.42\,mm, respectively. $r_{1}$ is the effective radius of the MCP measurement cable at the MCP joint, with a value of 16.25\,mm. $r_{g}$ denotes the radius of the gear connected to the encoder. $\theta_{e_m}$ and $\theta_{e_d}$ are the angles measured by the encoders at the MCP and DIP joints, respectively. $\Delta L_{1}'$, $\Delta L_{2}'$, and $\Delta L_{3}'$ represent the elongations of the DIP measurement cable at the MCP, PIP, and DIP joints, respectively, when the finger is fully flexed. The constants 0.230 and 0.989 in \eqref{equ:Θ2} are obtained experimentally \cite{6987303}.
By solving these kinematic equations, we obtain the operator's finger joint angles, which are subsequently mapped to the dexterous hand for real-time control.

\subsection{Force-feedback Cable Following Calculation}
\label{subsec:force jisuan formula}

In our force feedback system, the servo must actively manage the force-feedback cable to follow the operator's natural finger movements. For instance, when the operator's finger flexes, the cable path along the dorsal side elongates, requiring the servo to pay out the cable to maintain a constant tension. In contrast, when the finger extends, the cable path shortens, and the servo must retract the cable to take up slack. An accurate calculation of this required cable displacement is essential for transparent operation. To enable this motion following, the system must continuously calculate the target displacement of the force-feedback cable. This displacement is calculated directly from the posture of the finger, specifically, the joint angles derived in the previous section. The geometric model for this calculation, illustrated in Fig. \ref{fig:shouzhijiesuan}(b), is based on the coordinate transformations detailed below:

\begin{equation}
\begin{aligned}
\theta_{s} = \frac{L_a}{r_{s}}
\end{aligned}
\label{equ:xiata_s}
\end{equation}

\begin{equation}
\begin{split}
L_a = & \left| P_{A_1}' - P_{O_1}' \right| + \left| P_{A_2}' - P_{A_1}' \right| + \left| P_{B_1}' - P_{A_2}' \right| \\
& + \left| P_{B_2}' - P_{B_1}' \right| + \left| P_{C_1}' - P_{B_2}' \right| + \left| P_{C_2}' - P_{C_1}' \right| + \sigma
\end{split}
\label{equ:L_a}
\end{equation}

\begin{equation}
\begin{aligned}
P_{A_1}' &= T_{01} P_{A_1}, \\
P_{A_2}' &= T_{01} P_{A_2}, \\
P_{B_1}' &= T_{01} T_{12} P_{B_1}, \\
P_{B_2}' &= T_{01} T_{12} P_{B_2}, \\
P_{C_1}' &= T_{01} T_{12} T_{23} P_{C_1}, \\
P_{C_2}' &= T_{01} T_{12} T_{23} P_{C_2}
\end{aligned}
\label{equ:transformations}
\end{equation}

\begin{equation}
T_{n-1,n} = \begin{bmatrix}
\cos\theta_n & -\sin\theta_n & 0 & -l_{n-1} \\
\sin\theta_n & \cos\theta_n & 0 & 0 \\
0 & 0 & 1 & 0 \\
0 & 0 & 0 & 1
\end{bmatrix}
\label{equ:T_matrix}
\end{equation}
where $\theta_{s}$ denotes the rotation angle of the servo and $r_{s}$ represents the radius of the output flange.
$L_a$ represents the length of the force-feedback cable with respect to the base coordinate origin $O$. $P_{A_1}$, $P_{A_2}$, $P_{B_1}$, $P_{B_2}$, $P_{C_1}$, and $P_{C_2}$ denote the vectors of the corresponding points with respect to the local coordinate systems at points A, B, and C, respectively. 
$l_0$, $l_1$, $l_2$, $l_3$ are the lengths of the segments between the base coordinate origin $O$, the points $A$, $B$, $C$, and $D$, respectively.
Their lengths are 35.71\,mm, 44.33\,mm, 24.21\,mm, and 23.51\,mm, respectively. $P_{A_1}'$, $P_{A_2}'$, $P_{B_1}'$, $P_{B_2}'$, $P_{C_1}'$, and $P_{C_2}'$ are the vectors of the corresponding points with respect to $O$. $\sigma$ is a cable slack variable, typically set to 2\,mm. $T_{n-1,n}$ is the transformation matrix from the coordinate frame $n-1$ to the frame $n$. Angle $\theta_n$ and length $l_{n-1}$ are defined according to the kinematic model of the finger joints, where $\theta_n$ is obtained from Sec. \ref{subsec:encoder jisuan formula}. 
Through the transformations in \eqref{equ:transformations}, the local coordinates can be converted to global coordinates. Once the value of $L_a$ is obtained, $\theta_{s}$ can be calculated using \eqref{equ:xiata_s}, thereby enabling precise determination of the target force-feedback cable displacement.

\section{Experiments}
\label{sec:experiments}
In this section, we evaluate the performance of CDF-Glove.
The overall parameters of the CDF-Glove are presented in Table. \ref{tab:CDF-Glove Parameters}, and the cost is summarized in Table. \ref{tab:CDF-Glove Cost}.

\begin{figure}[t]
    \centering
    \includegraphics[width=\linewidth]{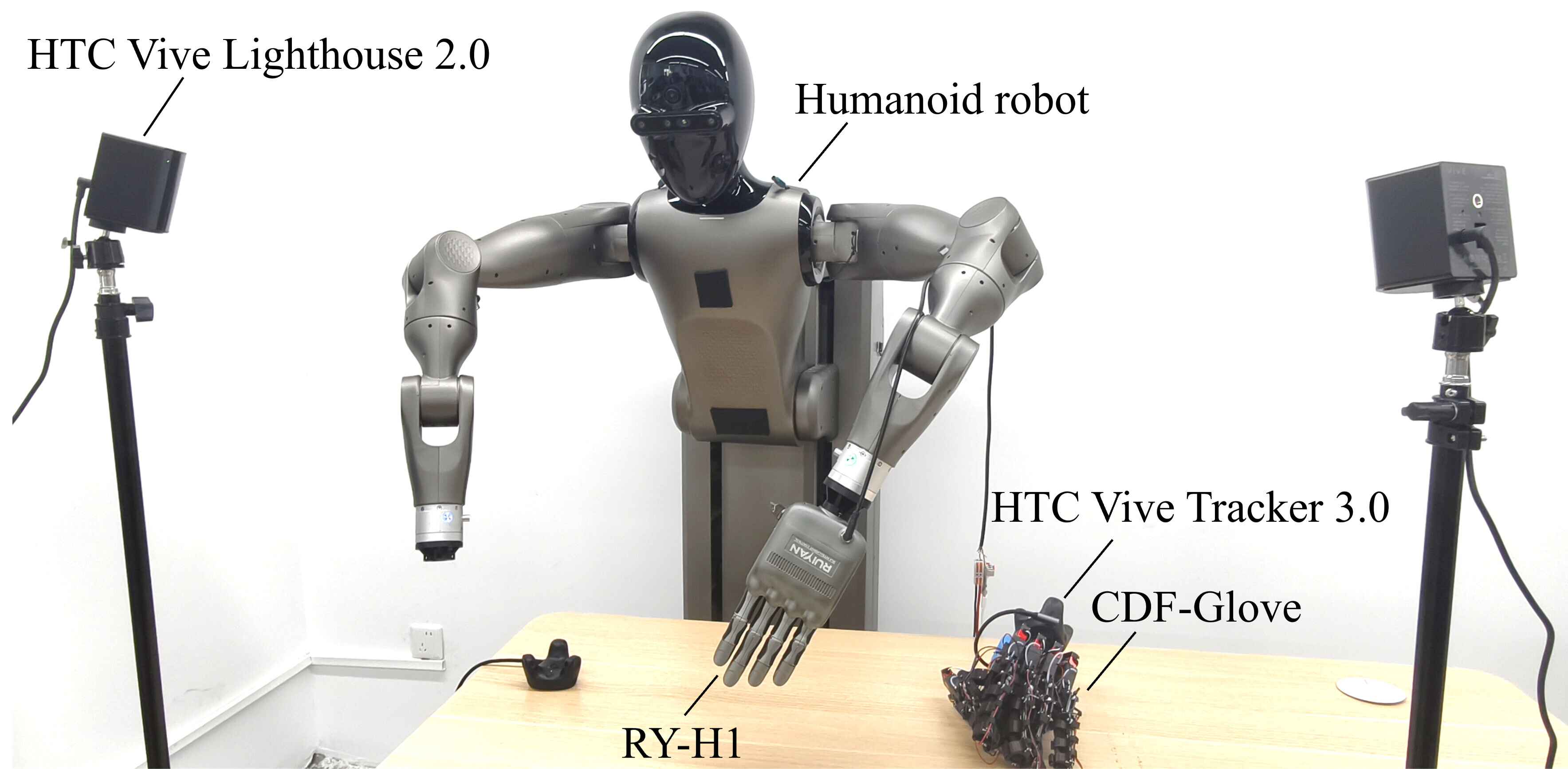}
    \caption{Overview of the complete experimental test system.}
    \label{fig:system}
    \vspace{-0.3cm}
\end{figure}

A series of experiments are conducted on a bimanual humanoid robot. As shown in Fig. \ref{fig:system}, the CDF-Glove is used to teleoperate the RY-H1 dexterous hand from \textit{RUIYAN}, which features 15 actively actuated DoF. The maximum data acquisition frequency of the CDF-Glove is approximately 100 Hz. The wrist tracking is accomplished using the \textit{HTC} Vive Tracker 3.0 and the \textit{HTC} Vive Lighthouse 2.0, with 6D pose estimation provided by \textit{SteamVR}.

\begin{table}[!htbp]
\centering
\caption{CDF-Glove Overall Performance Parameters}
\label{tab:CDF-Glove Parameters}
\begin{tabular}{@{}ll@{}}
\toprule
\textbf{Items} & \textbf{Parameters} \\ \midrule
Weight (including servos) & 0.49 kg \\
Force feedback latency to hand & 200 ms \\
Measurable DoF & 16 \\
Coupled computation DoF & 4 \\
Finger width & 25 mm \\
Communication protocol & RS485 \& Modbus @ 0.5Mbits/s \\
Maximum Power & 50 Watt @ 5 Volt \\ \bottomrule
\end{tabular}
\end{table}

\begin{table}[!htbp]
\centering
\caption{Cost of CDF-Glove and Others'}
\label{tab:CDF-Glove Cost}
\begin{tabular}{@{}cccc@{}}
\toprule
\textbf{Type} & \textbf{Item} & \textbf{Num} & \textbf{Price} \\ \midrule
\multirow{4}{*}{Electronic components} & Control board & 1 & \$90.2 \\ 
& Servos HLS3606M & 5 & \$107.1 \\
& LRAs & 5 & \$7.14 \\
& Encoders & 16 & \$4.13 \\ \midrule
\multirow{2}{*}{Non-standard parts} 
& 3D printed parts &  & \$14.28 \\
& Other materials & & \$3.03 \\ \midrule
\multirow{5}{*}{Other components} & Cable & 15 & \$8.65 \\
& Screws & & \$0.35 \\
& Gaskets & & \$0.09 \\
& PTFE sleeves & & \$0.11 \\ 
& Springs &  & \$4.08 \\ \midrule
\textbf{Total cost} & & & \$230.51 \\ \bottomrule

\midrule
DOGlove \cite{zhang2025doglove} & & & \$600 \\
GEX Series \cite{dong2025gexdemocratizingdexterityfullyactuated} & & & \$600 \\
\midrule

\end{tabular}
\end{table}

\subsection{CDF-Glove Performance Validation}
\label{subsec:CDF-Glove Performance Validation}
\subsubsection{CDF-Glove Basic Control Demonstration}
\label{subsubsec:CDF-Glove Basic Control Demonstration}
\begin{figure}[t]
    \centering
    \includegraphics[width=\linewidth]{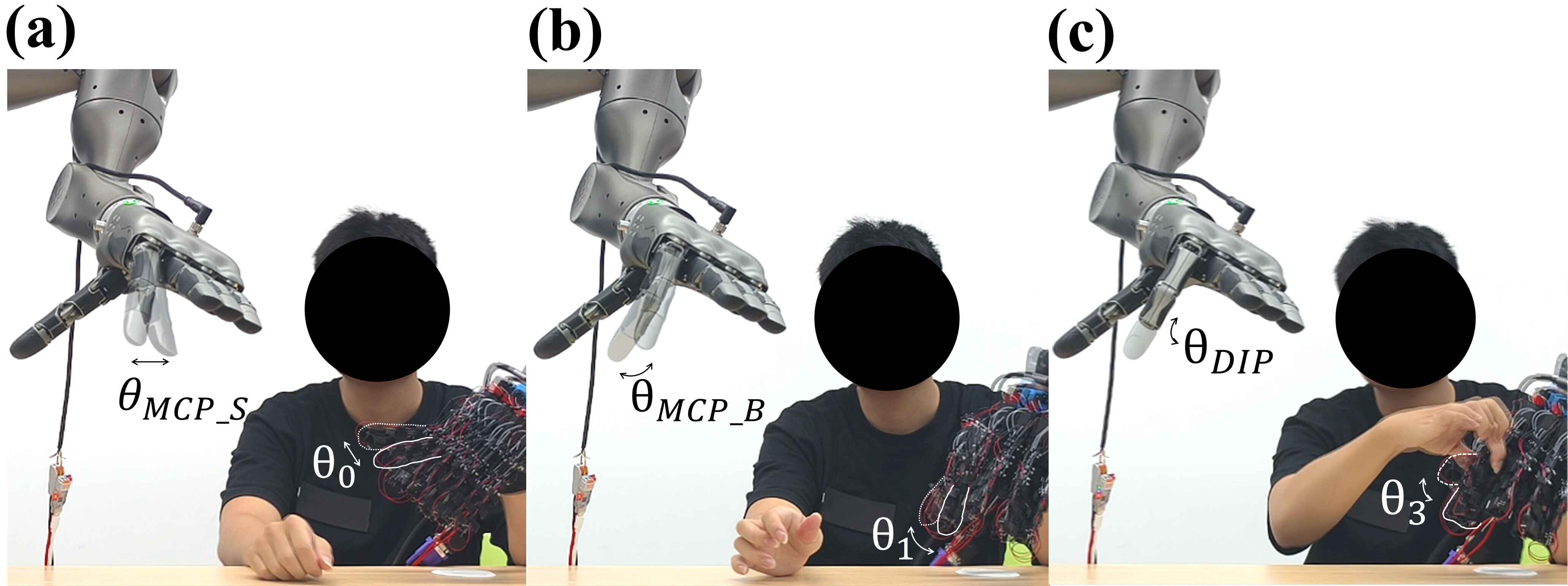}
    \caption{Basic movements of the 3-DoF finger joints. (a) MCP joint split. (b) MCP joint bend. (c) DIP joint flexion.}
    \label{fig:basic_control}
    \vspace{-0.3cm}
\end{figure}

To verify the effectiveness of teleoperated motion mapping with the CDF-Glove, a basic control demonstration experiment is conducted. In this experiment, the operator wears the CDF-Glove and controls the RY-H1 by performing the corresponding finger joint movements. 
Fig. \ref{fig:basic_control} illustrates the split motion of the MCP joint, the bend motion of the MCP joint, and the bend motion of the DIP joint, respectively.
The results demonstrate successful teleoperation. The RY-H1 hand visibly mirrors the movements of the operator's fingers. The system proves capable of mapping various gestures, from individual joint motions to complex grasping postures.

\subsubsection{CDF-Glove Force Feedback Performance}
\label{subsubsec:CDF-Glove Force Feedback Performance}
\begin{figure}[t]
    \centering
    \includegraphics[width=\linewidth]{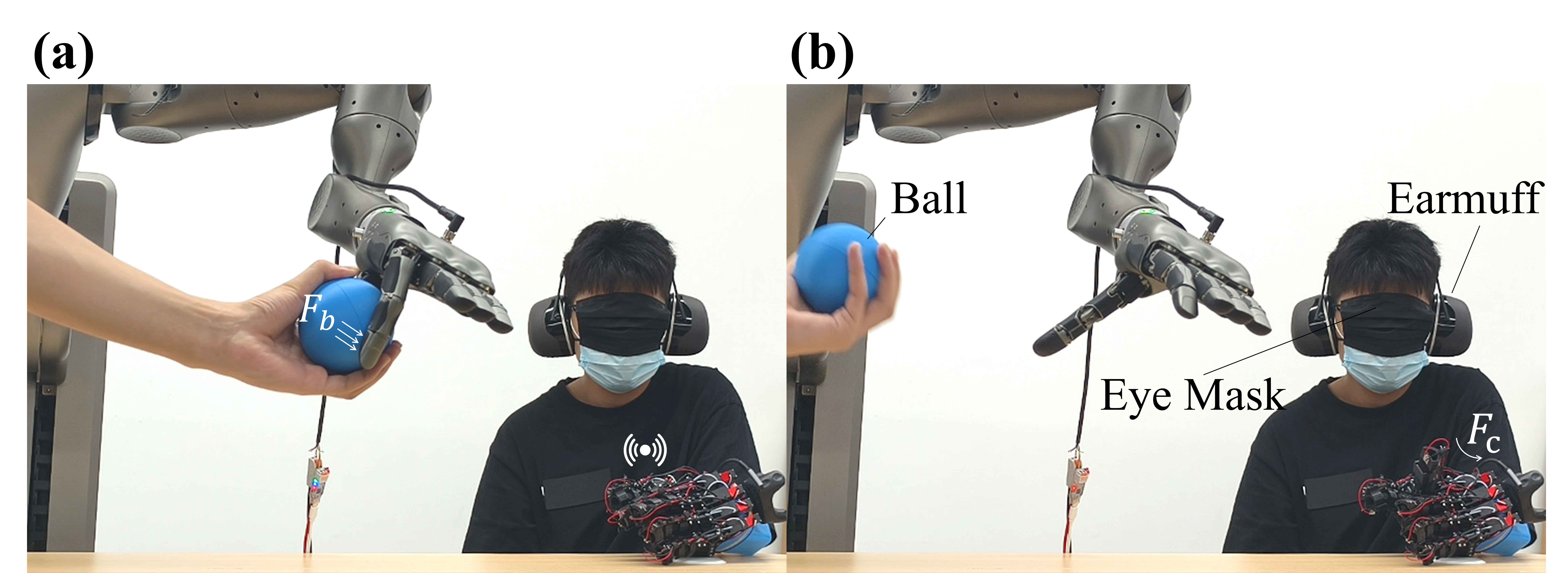}
    \caption{Schematic diagram of the overall haptic feedback process. (a) represents the vibration signal when force is applied to the index finger. (b) represents the generation of cable-driven feedback.}
    \label{fig:force_feedback}
    \vspace{-0.3cm}
\end{figure}

To validate the haptic feedback performance of the CDF-Glove, as described in Sec. \ref{subsec:Haptic and Force Feedback System} and Sec. \ref{subsec:force jisuan formula}, a dedicated experiment is conducted. 
The force signal is converted into the current signal of the RY-H1's motor, which is then used to simulate haptic feedback.
In the experiment, the index finger of RY-H1 is brought into contact with a rigid sphere, as shown in Fig. \ref{fig:force_feedback}. The contact force, measured by the DIP motor of the RY-H1, is denoted as $F _b$. This force signal is then used to trigger the LRA on the operator's corresponding fingertip, providing vibrotactile feedback.
When the applied force exceeds 1\,N (equivalent to 65\,mA), the force feedback system is activated, causing the cable to tighten and generate resistive feedback $F_{c}$, as shown in Fig. \ref{fig:force_feedback}(b). 

\begin{figure}[H]
    \centering
    \includegraphics[width=\linewidth]{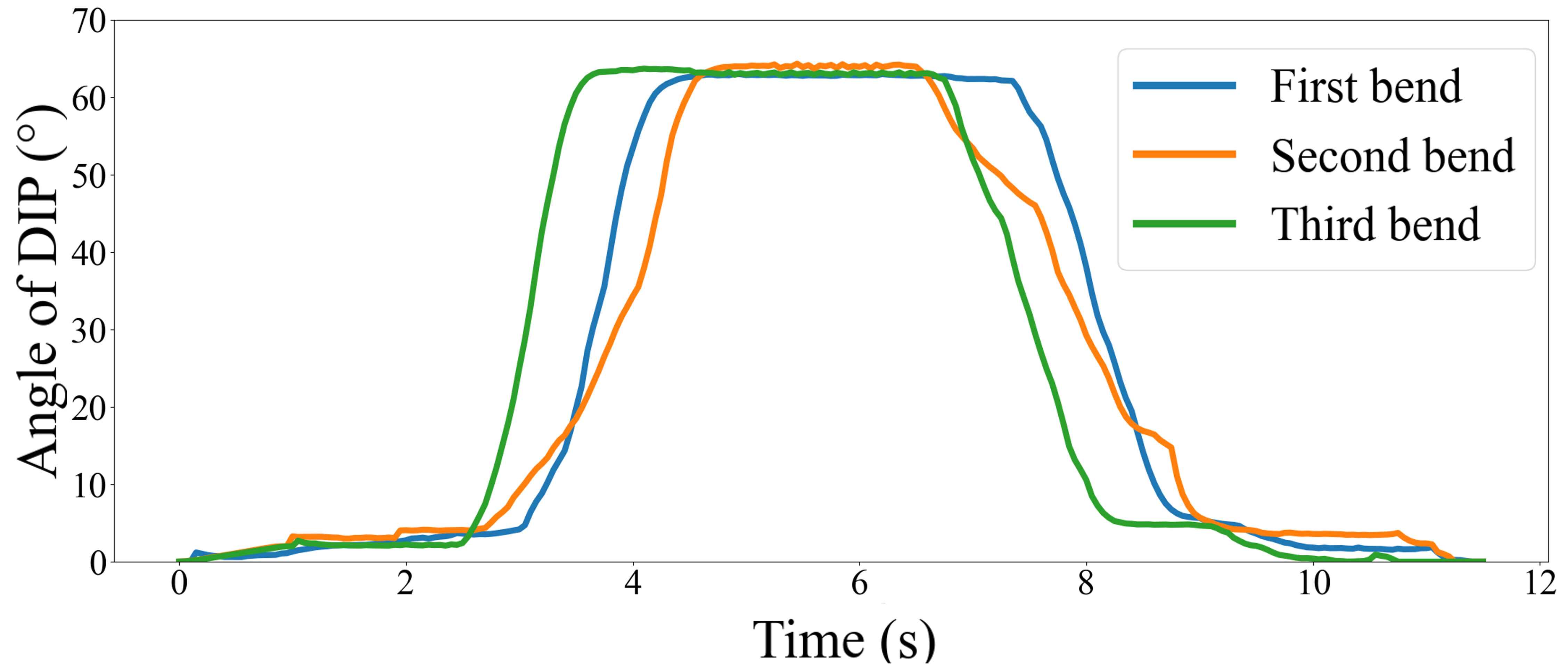}
    \caption{Results of three repeated positioning experiments for the index finger DIP joint.}
    \label{fig:retarget}
    \vspace{-0.3cm}
\end{figure}
\begin{figure}[t]
    \centering
    \includegraphics[width=\linewidth]{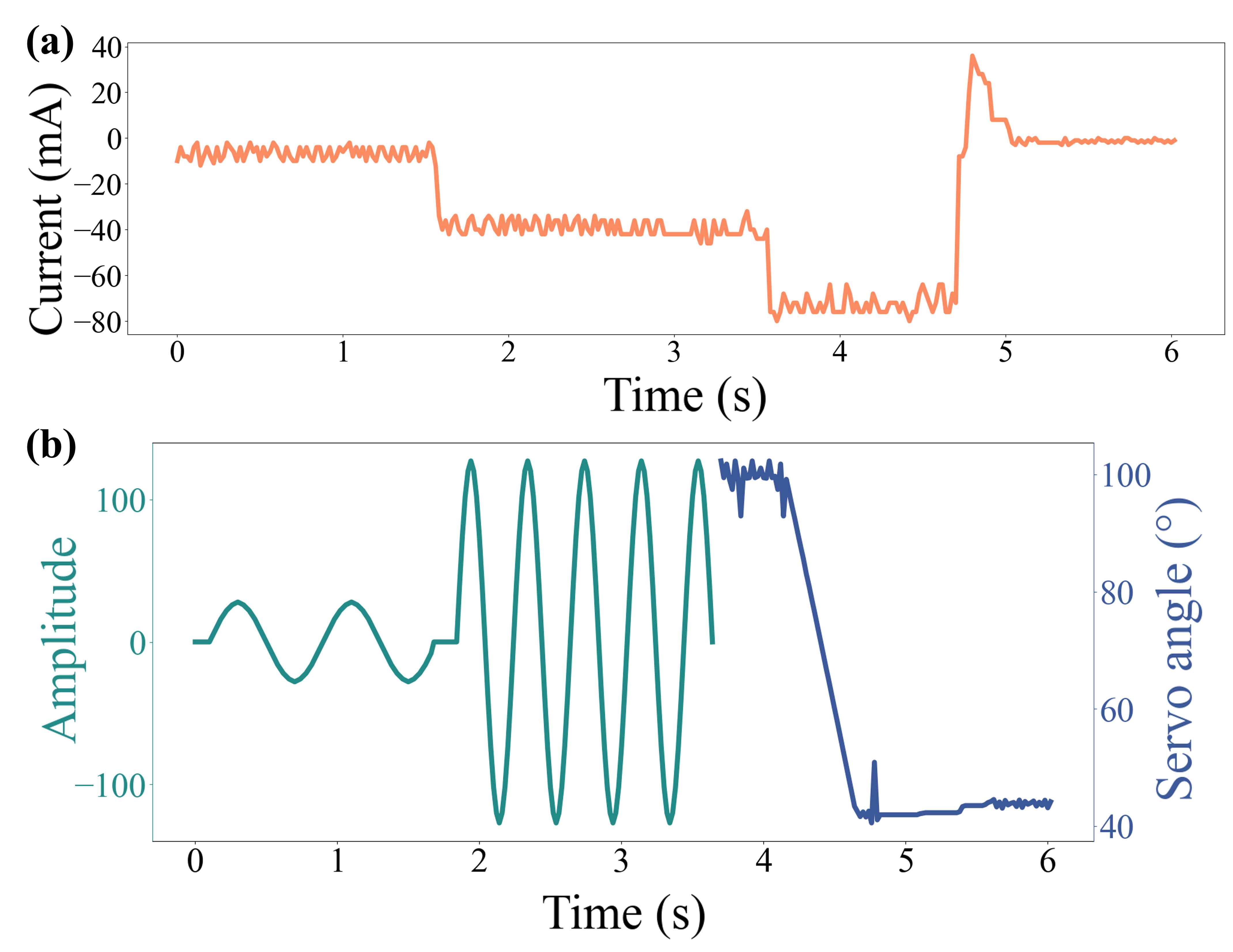}
    \caption{Haptic feedback processes information for the index finger. (a) shows the current variation curve of the RY-H1. (b) presents the waveform of the LRA and the servo angle, including mode switching.}
    \label{fig:lifankui}
    \vspace{-0.3cm}
\end{figure}

This force counteracts the operator's input, effectively halting further flexion and creating a pulling sensation back toward the finger's straighter, initial posture.
Fig. \ref{fig:lifankui} presents detailed information on the haptic feedback process for the index finger: Fig. \ref{fig:lifankui}(a) shows the current variation curve of RY-H1 during the experiment, while Fig. \ref{fig:lifankui}(b) displays the LRA waveform and the servo angle variation curve, including mode switching. The experimental results indicate that the response time for cable-driven force feedback, based on current information from the RY-H1, is nearly 200 ms.
The 200 ms force-feedback latency is primarily attributed to the serial communication bottleneck (RS485) and the mechanical response time of the cable-driven servos. While relatively high for high-speed interactions, it proved sufficient for the quasi-static manipulation tasks in our datasets.

\begin{figure}[t]
    \centering
    \includegraphics[width=\linewidth]{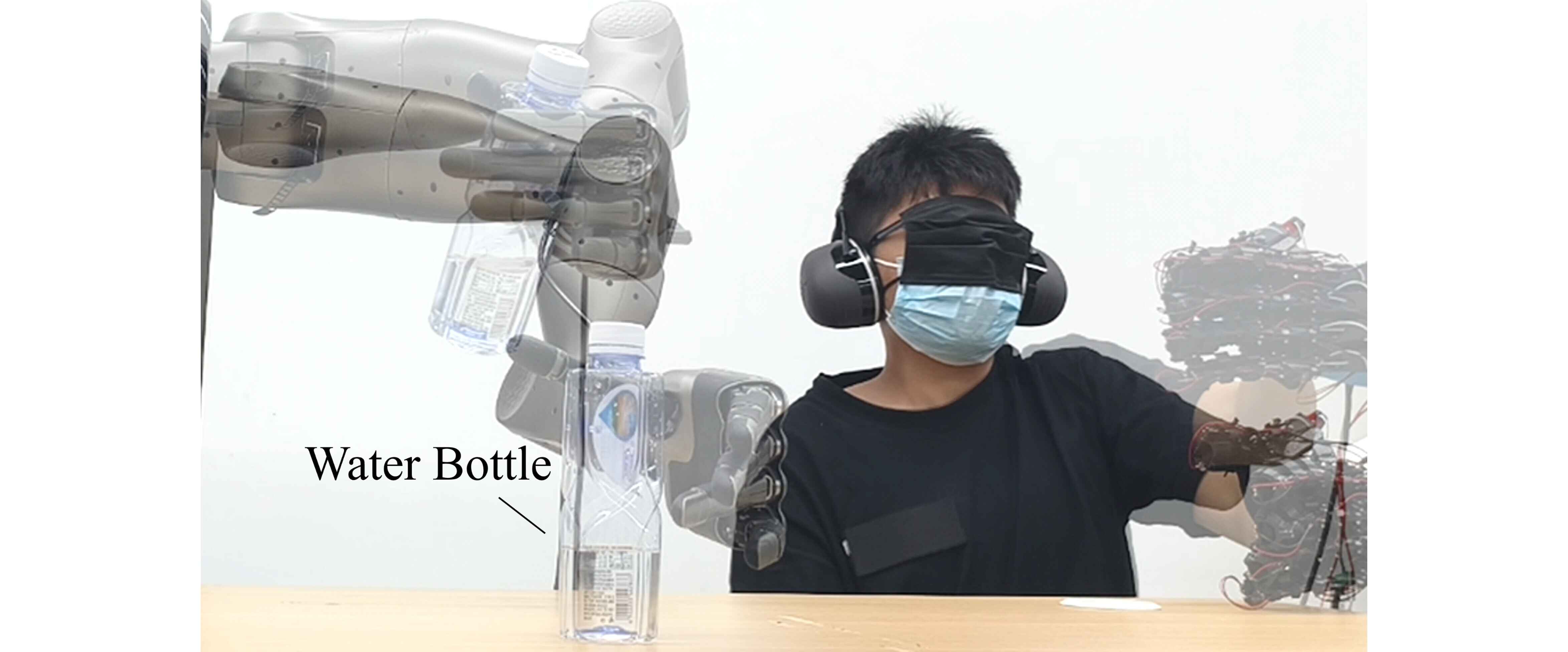}
    \caption{Bottle grasping experiment based on haptic feedback.}
    \label{fig:force_feedback_bottle}
    \vspace{-0.3cm}
\end{figure}

Finally, a bottle grasping experiment based on haptic feedback is conducted, as shown in Fig. \ref{fig:force_feedback_bottle}. The tests are performed under two conditions: with eye mask and earmuff, and without eye mask or earmuff. Each condition is further divided into trials with and without haptic feedback. The success rate of grasping a water bottle in these scenarios is summarized in Table. \ref{tab:bottle_Force_Feedback_Performance}, where haptic feedback is denoted as F and the use of eye mask and earmuff as E. 
Experimental results show that, under the eye mask and earmuff condition, the introduction of haptic feedback increases the success rate from 10\% to 50\%, representing a 4x improvement.
Similarly, in the absence of eye mask and earmuff, haptic feedback increases the success rate from 70\% to 90\%. 

\begin{table}[!htbp]
\centering
\caption{Performance for Grasping a Water Bottle}
\label{tab:bottle_Force_Feedback_Performance}
\begin{tabular}{@{}ccc@{}}
\toprule
\textbf{Condition} & \textbf{Success Rate} & \textbf{Average Completion Time (s)}\\ 
\midrule
$\checkmark F  \checkmark E  $   & \textbf {5/10} & \textbf {8.52}\\
$\times F  \checkmark E  $ & 1/10 & 18.30\\
$\checkmark F  \times E$ & \textbf {9/10}& \textbf {2.51}\\
$\times F  \times E$ & 7/10 & 3.11\\
\bottomrule
\end{tabular}
\end{table}
These findings validate the effectiveness and importance of the haptic feedback system in teleoperation tasks.

\begin{figure*}[!thbp]
    \centering
    \includegraphics[width=\linewidth]{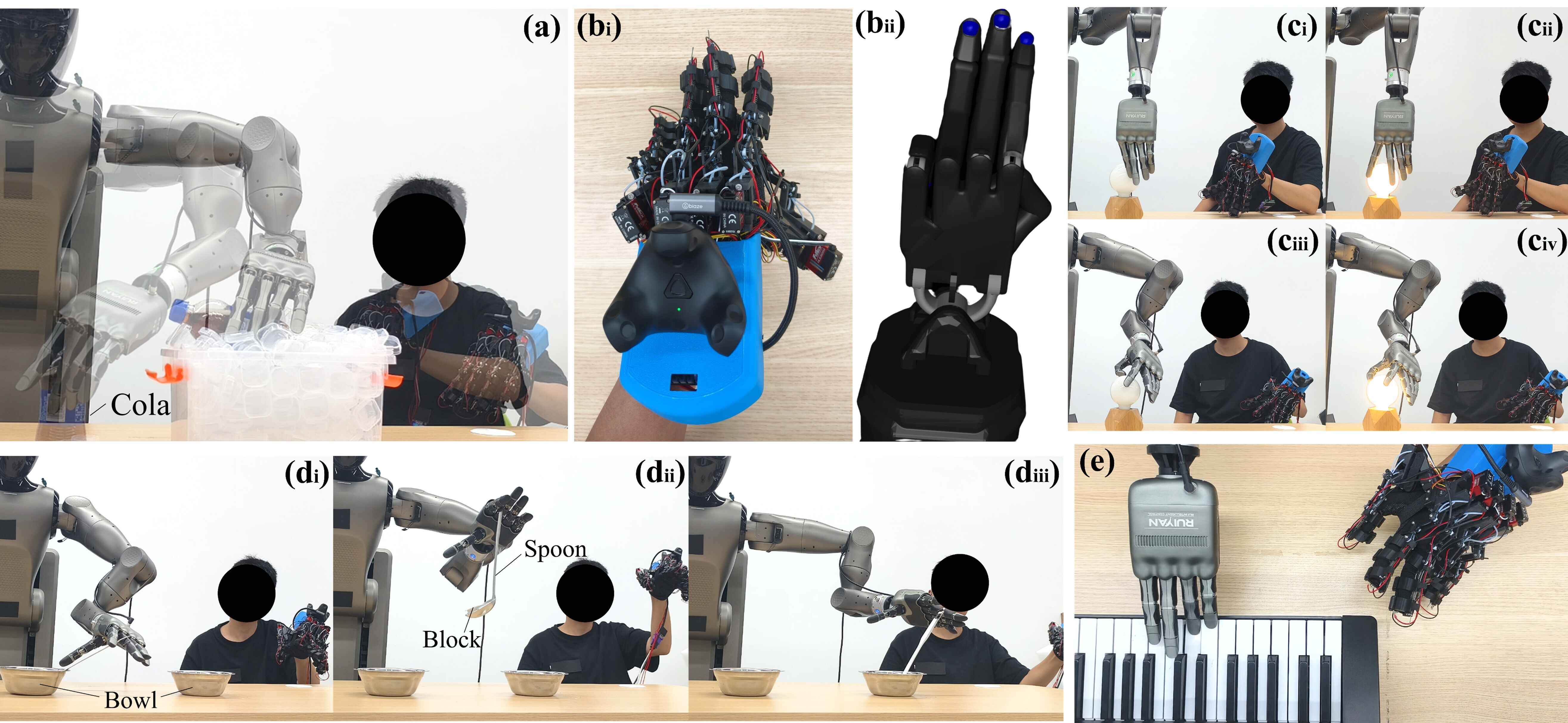}
    \caption{
    Demonstrations of various tasks using the CDF-Glove. (a) Cola transportation task: removing a cola bottle from a box and placing it upright on the table. (b)
    Gesture demonstration: performing hand gestures based on the SDH model. (c) Manipulation task: rotating a light bulb with one or two fingers to turn it on. (d) Transport task: moving a spoon and the block on top of it from one bowl to another.  (e) Manipulation task: playing the piano.}
    \label{fig:manipulation}
    \vspace{-0.3cm}
\end{figure*}

\subsubsection{CDF-Glove Repeated Positioning Performance}
\label{subsubsec:CDF-Glove Repeated Positioning Performance}
To evaluate the fingertip repeated positioning of the CDF-Glove, we perform experiments on the index finger DIP joint. The index finger is first reset to the zero position. The finger then flexes to a specific angle (so that the fingertip touches an object), and subsequently returns to the zero position. This process repeats three times. We calculate the mean and standard deviation of the contact angle, resulting in an average value of 63.15°, with standard deviations of 0.29°, respectively. As shown in Fig. \ref{fig:retarget}, the standard deviation of repeated positioning for the index finger distal joint (DIP) using CDF-Glove is less than 0.4°, indicating high accuracy and stability. Similar tests conducted on the MCP and PIP joints of the index and middle fingers show consistent stability, with standard deviations remaining below 0.4°.

\subsection{CDF-Glove System Modeling Validation}
\label{subsec:CDF-Glove System Modeling Validation}
To validate the effectiveness of the finger joint kinematic model proposed in Sec. \ref{subsec:encoder jisuan formula}, experiments are conducted using three dexterous hands with different DoF: the 6-DoF RY-H2, the 15-DoF RY-H1, and the 24-DoF Shadow Dexterous Hand (SDH). The operation tasks for RY-H2 are verified in Sec. \ref{subsec:Bimanual CDF-Glove Teleoperation Data Collection}. For RY-H1, validation is performed through grasping, transporting, and manipulation tasks, as illustrated in Fig. \ref{fig:manipulation}. For example, a cola bottle is grasped and placed upright on the table, requiring precise haptic feedback to control the hand posture and rotate the bottle. A spoon with a block is transported from one bowl to another, necessitating stable and accurate finger coordination throughout the process. Additional tasks, such as playing piano and rotating the light bulb (using single or dual fingers), further demonstrate the teleoperation capabilities of the CDF-Glove. For SDH, validation is performed in simulation through gesture demonstrations. Each finger has four active DoF, allowing both DIP and PIP joints to be actively controlled, thus effectively verifying the correctness of (\ref{equ:Θ2}). Experimental results show that the CDF-Glove is based on the kinematic model proposed in Sec. \ref{subsec:encoder jisuan formula}, enables effective motion control of varying dexterous hands.

\subsection{Bimanual CDF-Glove Teleoperation Data Collection}
\label{subsec:Bimanual CDF-Glove Teleoperation Data Collection}
To verify the high-quality data acquisition capability of the CDF-Glove, we conduct bimanual teleoperation data collection experiments (use \textit{AGIBOT}). Bimanual tasks place significant demands on the coordination and dexterity of both hands and arms, emphasizing collaborative performance and fully demonstrating the CDF-Glove's ability to capture data in complex scenarios. Furthermore, successful completion of these tasks supplements the experiments on the RY-H2 described in Sec. \ref{subsec:CDF-Glove System Modeling Validation}.
The operator wears two CDF-Glove devices and uses two trackers to track wrist positions and orientations, enabling the teleoperation of two RY-H2 to perform tasks such as stacking paper cups and transferring a plastic film roll, as illustrated in Fig. \ref{fig:shuangbi}(a) and (c).
For fingers other than the thumb, we map the DIP joint to the RY-H2. For thumb, both the IP and TM(S) joints are mapped to the RY-H2.
Throughout the experiments, the operator flexibly adjusts the postures of the finger and applies tactile sensation according to the specific requirements of each task. The entire data acquisition pipeline operates at approximately 30\,Hz. The data collected include 14 joint angles for both arms, one vertical displacement measurement, six joint angles for each RY-H2, and image information. The image data are acquired using a \textit{Realsense} D455 camera and two \textit{Realsense} D405 cameras.

\subsection{Diffusion Policy on Collected Data}
\label{subsec:Diffusion Policy on Collected Data}
To evaluate the quality of the data collected using the CDF-Glove, we employ the IL method known as Diffusion Policy \cite{chi2024diffusionpolicy}, which has demonstrated impressive performance in various robotic manipulation tasks. 
For each task, we train the Diffusion Policy models on two datasets: one collected using the CDF-Glove (see Sec. \ref{subsec:Bimanual CDF-Glove Teleoperation Data Collection}) and the other collected through kinematic teaching (KT).
Both datasets consist of 200 teleoperated demonstrations for the cup stacking task and another 200 for the plastic film roll transfer task, totaling 400 demonstrations per dataset.
The training procedure for both models is identical, each is trained for 1000 epochs.

The validation results of the Diffusion Policy models for both tasks and datasets are presented in Table. \ref{tab:Success Rate and Completion Time for the Two Tasks}. 
\begin{figure}[t]
    \centering
    \includegraphics[width=\linewidth]{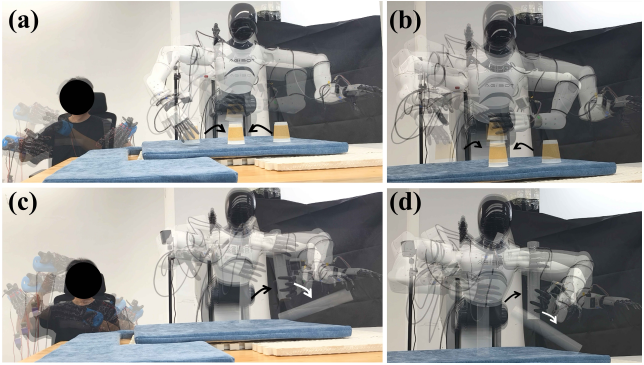}
        \caption{Bimanual data collection. (a) Teleoperate cup stacking. (b) Cup stacking using the trained model. (c) Teleoperate plastic film roll transfer. (d) Plastic film roll transfer using the trained model.}
    \label{fig:shuangbi}
    \vspace{-0.3cm}
\end{figure}
The execution of the trained models on the two tasks is illustrated in Fig. \ref{fig:shuangbi}(b) and (d).
It can be observed that the dataset collected with CDF-Glove is of higher quality than the KT dataset: it increases the average success by 55\% and shortens the completion time by $\approx$15.2 seconds (a 47.2\% relative reduction).

\begin{table}[t]
\centering
\caption{Success Rate and Completion Time for the Two Tasks}
\label{tab:Success Rate and Completion Time for the Two Tasks}
\begin{tabular}{@{}ccc@{}}
\toprule
\textbf{Condition} & \textbf{Success Rate} & \textbf{Average Completion Time (s)}\\ 
\midrule
CDF (Cup Stacking)   & \textbf{10/10} & \textbf{14.74}\\
KT (Cup Stacking)    & 4/10 & 28.91\\
CDF (Film Roll Transfer)   & \textbf{8/10} & \textbf{19.49}\\
KT (Film Roll Transfer)    & 3/10 & 35.70\\
\bottomrule
\end{tabular}
\end{table}

\section{Conclusion and Future Work}
\label{sec:conclusion}
This work presents the CDF-Glove: a lightweight, high-degree-of-freedom, cable-driven force-feedback glove.  
The device attains 0.4° repeatability in distal-joint positioning and delivers force feedback within 200 ms.  
Comprehensive kinematic models of finger-joint angles and cable-tracking are derived and experimentally verified.  
Compared with KT, bimanual data acquisition with the CDF-Glove followed by IL increases task success by 55\% and reduces mean completion time by $\approx$ 15.2 seconds (a 47.2\% relative reduction). 
In particular, the CDF-Glove costs $\approx$ \$230. 
Current limitations include restricted force output, lack of complex haptic feedback, encoder signal drift and zero-offset fluctuation, as well as PCB-induced control latency.  
Ongoing work will focus on enhancing the force output, optimizing the PCB design to minimize latency, and integrating cable-driven control strategies\cite{11343246} to further reduce tracking errors.


\bibliographystyle{IEEEtran}
\bibliography{bib/bibliography}

@article{ARGALL2009469,
title = {A survey of robot learning from demonstration},
journal = {Robotics and Autonomous Systems},
volume = {57},
number = {5},
pages = {469-483},
year = {2009},
issn = {0921-8890},
doi = {https://doi.org/10.1016/j.robot.2008.10.024},
author = {Brenna D. Argall and Sonia Chernova and Manuela Veloso and Brett Browning}
}

@inproceedings{qin2023anyteleop,
  title     = {AnyTeleop: A General Vision-Based Dexterous Robot Arm-Hand Teleoperation System},
  author    = {Qin, Yuzhe and Yang, Wei and Huang, Binghao and Van Wyk, Karl and Su, Hao and Wang, Xiaolong and Chao, Yu-Wei and Fox, Dieter},
  booktitle = {Robotics: Science and Systems},
  year      = {2023}
}

@inproceedings{ben2024homie,
  title={HOMIE: Humanoid Loco-Manipulation with Isomorphic Exoskeleton Cockpit},
  author={Ben, Qingwei and Jia, Feiyu and Zeng, Jia and Dong, Junting and Lin, Dahua and Pang, Jiangmiao},
  booktitle={RSS 2025 Workshop on Whole-body Control and Bimanual Manipulation: Applications in Humanoids and Beyond},
  year={2025}
}

@inproceedings{wang2024dexcap,
  title={DexCap: Scalable and Portable Mocap Data Collection System for Dexterous Manipulation},
  author={Wang, Chen and Shi, Haochen and Wang, Weizhuo and Zhang, Ruohan and Fei-Fei, Li and Liu, Karen},
  booktitle={RSS 2024 Workshop: Data Generation for Robotics},
  year={2024},
}

@article{zhao2023learningfinegrainedbimanualmanipulation,
  title={Learning Fine-Grained Bimanual Manipulation with Low-Cost Hardware},
  author={Zhao, Tony and Kumar, Vikash and Levine, Sergey and Finn, Chelsea},
  journal={Robotics: Science and Systems XIX},
  year={2023},
  publisher={Robotics: Science and Systems Foundation}
}

@inproceedings{fang2025airexo2scalinggeneralizablerobotic,
  title={AirExo-2: Scaling up Generalizable Robotic Imitation Learning with Low-Cost Exoskeletons},
  author={Fang, Hongjie and Wang, Chenxi and Wang, Yiming and Chen, Jingjing and Xia, Shangning and Lv, Jun and He, Zihao and Yi, Xiyan and Guo, Yunhan and Zhan, Xinyu and others},
  booktitle={7th Robot Learning Workshop: Towards Robots with Human-Level Abilities},
  year={2025}
}

@article{zhang2025doglove,
  title={DOGlove: Dexterous Manipulation with a Low-Cost Open-Source Haptic Force Feedback Glove},
  author={Zhang, Han and Hu, Songbo and Yuan, Zhecheng and Xu, Huazhe},
  journal={arXiv preprint arXiv:2502.07730},
  year={2025}
}

@misc{dong2025gexdemocratizingdexterityfullyactuated,
      title={GEX: Democratizing Dexterity with Fully-Actuated Dexterous Hand and Exoskeleton Glove}, 
      author={Yunlong Dong and Xing Liu and Jun Wan and Zelin Deng},
      year={2025},
      eprint={2506.04982},
      archivePrefix={arXiv},
      primaryClass={cs.RO},
}

@INPROCEEDINGS{8206575,
  author={Liu, Hangxin and Xie, Xu and Millar, Matt and Edmonds, Mark and Gao, Feng and Zhu, Yixin and Santos, Veronica J. and Rothrock, Brandon and Zhu, Song-Chun},
  booktitle={2017 IEEE/RSJ International Conference on Intelligent Robots and Systems (IROS)}, 
  title={A glove-based system for studying hand-object manipulation via joint pose and force sensing}, 
  year={2017},
  volume={},
  number={},
  pages={6617-6624},
  doi={10.1109/IROS.2017.8206575}}

@INPROCEEDINGS{8794230,
  author={Liu, Hangxin and Zhang, Zhenliang and Xie, Xu and Zhu, Yixin and Liu, Yue and Wang, Yongtian and Zhu, Song-Chun},
  booktitle={2019 International Conference on Robotics and Automation (ICRA)}, 
  title={High-Fidelity Grasping in Virtual Reality using a Glove-based System}, 
  year={2019},
  volume={},
  number={},
  pages={5180-5186},
  doi={10.1109/ICRA.2019.8794230}}

@inproceedings{naughton2024respilotteleoperatedfingergaiting,
  title={ResPilot: Teleoperated Finger Gaiting via Gaussian Process Residual Learning},
  author={Naughton, Patrick and Cui, Jinda and Patel, Karankumar and Iba, Soshi},
  booktitle={Conference on Robot Learning},
  pages={4410--4424},
  year={2025},
  organization={PMLR}
}

@article{
doi:10.1126/scirobotics.adn3802,
author = {Shinichi Furuya  and Takanori Oku  and Hayato Nishioka  and Masato Hirano },
title = {Surmounting the ceiling effect of motor expertise by novel sensory experience with a hand exoskeleton},
journal = {Science Robotics},
volume = {10},
number = {98},
pages = {eadn3802},
year = {2025},
doi = {10.1126/scirobotics.adn3802}}

@INPROCEEDINGS{6569265,
  author={Popescu, Nirvana and Popescu, Decebal and Ivanescu, Mircea and Popescu, Dorin and Vladu, Cristian and Berceanu, Cosmin and Poboroniuc, Marian},
  booktitle={2013 19th International Conference on Control Systems and Computer Science}, 
  title={Exoskeleton Design of an Intelligent Haptic Robotic Glove}, 
  year={2013},
  volume={},
  number={},
  pages={196-202},
  doi={10.1109/CSCS.2013.21}}

@INPROCEEDINGS{7759176,
  author={Sarakoglou, Ioannis and Brygo, Anais and Mazzanti, Dario and Hernandez, Nadia Garcia and Caldwell, Darwin G. and Tsagarakis, Nikos G.},
  booktitle={2016 IEEE/RSJ International Conference on Intelligent Robots and Systems (IROS)}, 
  title={HEXOTRAC: A highly under-actuated hand exoskeleton for finger tracking and force feedback}, 
  year={2016},
  volume={},
  number={},
  pages={1033-1040},
  doi={10.1109/IROS.2016.7759176}}

@ARTICLE{8918345,
  author={Wang, Daoming and Wang, Yakun and Pang, Jiawei and Wang, Zhengyu and Zi, Bin},
  journal={IEEE Access}, 
  title={Development and Control of an MR Brake-Based Passive Force Feedback Data Glove}, 
  year={2019},
  volume={7},
  number={},
  pages={172477-172488},
  doi={10.1109/ACCESS.2019.2956954}}

@ARTICLE{9090275,
  author={Takahashi, Nobuhiro and Furuya, Shinichi and Koike, Hideki},
  journal={IEEE Transactions on Haptics}, 
  title={Soft Exoskeleton Glove with Human Anatomical Architecture: Production of Dexterous Finger Movements and Skillful Piano Performance}, 
  year={2020},
  volume={13},
  number={4},
  pages={679-690},
  doi={10.1109/TOH.2020.2993445}}

@article{CERULO201775,
title = {Teleoperation of the SCHUNK S5FH under-actuated anthropomorphic hand using human hand motion tracking},
journal = {Robotics and Autonomous Systems},
volume = {89},
pages = {75-84},
year = {2017},
issn = {0921-8890},
doi = {https://doi.org/10.1016/j.robot.2016.12.004},
author = {Ilaria Cerulo and Fanny Ficuciello and Vincenzo Lippiello and Bruno Siciliano}
}

@ARTICLE{6987303,
  author={Park, Yeongyu and Lee, Jeongsoo and Bae, Joonbum},
  journal={IEEE Transactions on Industrial Informatics}, 
  title={Development of a Wearable Sensing Glove for Measuring the Motion of Fingers Using Linear Potentiometers and Flexible Wires}, 
  year={2015},
  volume={11},
  number={1},
  pages={198-206},
  doi={10.1109/TII.2014.2381932}}

@incollection{RADULOVIC20241001,
title = {PTFE (polytetrafluoroethylene; Teflon®)},
editor = {Philip Wexler},
booktitle = {Encyclopedia of Toxicology (Fourth Edition)},
publisher = {Academic Press},
edition = {Fourth Edition},
address = {Oxford},
pages = {1001-1006},
year = {2024},
isbn = {978-0-323-85434-4},
doi = {https://doi.org/10.1016/B978-0-12-824315-2.00270-0},
author = {Louis L. Radulovic and Zbigniew W. Wojcinski}
}

@article{chi2024diffusionpolicy,
	author = {Cheng Chi and Zhenjia Xu and Siyuan Feng and Eric Cousineau and Yilun Du and Benjamin Burchfiel and Russ Tedrake and Shuran Song},
	title ={Diffusion Policy: Visuomotor Policy Learning via Action Diffusion},
	journal = {The International Journal of Robotics Research},
	year = {2024},
}

@article{JIANG2026103923,
title = {Prompt2Act: Mapping prompts into sequence of robotic actions with large foundation models},
journal = {Information Fusion},
volume = {127},
pages = {103923},
year = {2026},
issn = {1566-2535},
doi = {https://doi.org/10.1016/j.inffus.2025.103923},
url = {https://www.sciencedirect.com/science/article/pii/S1566253525009856},
author = {Maowei Jiang and Qi Wang and Hongfeng Ai and Zhiyong Dong and Yusong Hu and Ao Liang and Yifan Wang and Ruiqi Li and Quangao Liu and Moquan Chen and Peter Búš and Long Zeng}
}

@INPROCEEDINGS{10160491,
  author={Zhang, Hao and Liang, Hongzhuo and Cong, Lin and Lyu, Jianzhi and Zeng, Long and Feng, Pingfa and Zhang, Jianwei},
  booktitle={2023 IEEE International Conference on Robotics and Automation (ICRA)}, 
  title={Reinforcement Learning Based Pushing and Grasping Objects from Ungraspable Poses}, 
  year={2023},
  volume={},
  number={},
  pages={3860-3866},
  doi={10.1109/ICRA48891.2023.10160491}}

@article{Zheng2025DemonstratingDD,
  title={Demonstrating DVS: Dynamic Virtual-Real Simulation Platform for Mobile Robotic Tasks},
  author={Zijie Zheng and Zeshun Li and Yunpeng Wang and Qinghongbing Xie and Long Zeng},
  journal={ArXiv},
  year={2025},
  volume={abs/2504.18944},
  url={https://api.semanticscholar.org/CorpusID:278165721}
}

@INPROCEEDINGS{11343246,
  author={Liang, Huayue and Chen, Yanbo and Cheng, Hongyang and Yu, Yanzhao and Li, Shoujie and Tan, Junbo and Wang, Xueqian and Zeng, Long},
  booktitle={2025 IEEE International Conference on Systems, Man, and Cybernetics (SMC)}, 
  title={Data-Driven MPC with Data Selection for Flexible Cable-Driven Robotic Arms}, 
  year={2025},
  volume={},
  number={},
  pages={5742-5749},
  doi={10.1109/SMC58881.2025.11343246}}

@article{zhong2025dexgraspvla,
  title={Dexgraspvla: A vision-language-action framework towards general dexterous grasping},
  author={Zhong, Yifan and Huang, Xuchuan and Li, Ruochong and Zhang, Ceyao and Chen, Zhang and Guan, Tianrui and Zeng, Fanlian and Lui, Ka Num and Ye, Yuyao and Liang, Yitao and others},
  journal={arXiv preprint arXiv:2502.20900},
  year={2025}
}
\end{document}